\documentclass[letterpaper]{article} % DO NOT CHANGE THIS
\usepackage{aaai2027} % DO NOT CHANGE THIS
\usepackage[hyphens]{url} % DO NOT CHANGE THIS
\usepackage{graphicx} % DO NOT CHANGE THIS
\usepackage{natbib} % DO NOT CHANGE THIS AND DO NOT ADD ANY OPTIONS TO IT
\usepackage{caption} % DO NOT CHANGE THIS AND DO NOT ADD ANY OPTIONS TO IT
\usepackage{amsmath,amssymb,amsfonts,amsthm}
\usepackage{booktabs}
\usepackage{multirow}
\usepackage{algorithm}
\usepackage{algpseudocode}
\usepackage{tikz}
\usetikzlibrary{arrows.meta,positioning,shapes.geometric,fit,calc,decorations.pathreplacing,backgrounds}
\usepackage{xcolor}
\usepackage{xspace}

\usepackage{enumitem}

\newcommand{\clawloop}{\textsc{ClawLoop}\xspace}
\newcommand{\taskforge}{\textsc{TaskForge}\xspace}
\newcommand{\ie}{\emph{i.e.}\xspace}
\newcommand{\eg}{\emph{e.g.}\xspace}

\definecolor{phase1}{HTML}{4E79A7}
\definecolor{phase2}{HTML}{F28E2B}
\definecolor{phase3}{HTML}{E15759}
\definecolor{phase4}{HTML}{76B7B2}
\definecolor{loopbg}{HTML}{F5F5F5}

\title{ClawLoop: Self-Improving Agents via Harness-Native Training Loops}

\author{
    Hansi Yang\textsuperscript{1}\equalcontrib,
    Linji Hao\textsuperscript{2}\equalcontrib,
    Hongyan Hao\textsuperscript{1},
    An Wang\textsuperscript{1},
    Han Wu\textsuperscript{2}\corresponding,
    Qi Gu\textsuperscript{1}\corresponding,
    Xunliang Cai\textsuperscript{1}
}
\affiliations{
    \textsuperscript{1}Meituan, China \quad
    \textsuperscript{2} Peking University\\
    \texttt{yanghansi@meituan.com} \quad 
    \texttt{haolinji26@stu.pku.edu.cn} \quad \texttt{guqi03@meituan.com}
}
\AddToHookNext{shipout/foreground}{%
    \begin{tikzpicture}[remember picture, overlay]
        \node[
            anchor=north west,
            inner sep=0pt,
            xshift=0.75in,
            yshift=-0.40in
        ] at (current page.north west) {%
            \begin{minipage}{\textwidth}
                \raggedright
                \includegraphics[width=3.6cm]
                    {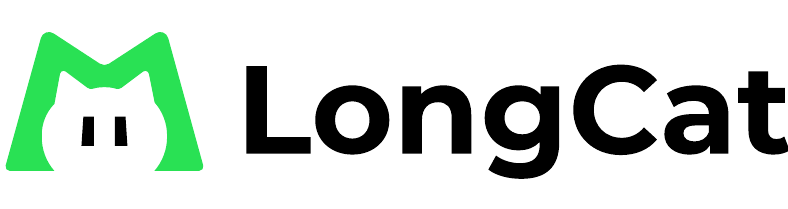}\par
                \vspace{3pt}
                \hrule height 0.4pt
            \end{minipage}%
        };
    \end{tikzpicture}%
}
\nocopyright
\begin{document}
\maketitle

% ======================================================================
%  ABSTRACT
% ======================================================================
\begin{abstract}
The rapid advancement of Large Language Models (LLMs) has driven their deployment toward general-purpose harness environments, where LLM-based agents operate in sandbox workspaces and interact directly with file systems and shells.
While numerous post-training pipelines exist for tool-use agents, their counterparts for harness-based agents remain scarce and are predominantly one-shot and teacher-dependent: a stronger external model synthesizes tasks and collects trajectories, the target agent is trained once, and evaluation results are consumed by human researchers rather than fed back into the loop.
Such pipelines are inherently bounded by the capability of the external model and the scope of human prior, leaving no room for self-improvement.
To address these limitations, we propose \clawloop{}, a framework that enables self-improving without any external teacher model and with minimal prior dependencies.
\clawloop{} composes three phases into a self-sustaining cycle: 
\emph{Task Synthesis}, which generates a diverse pool of training tasks via two complementary construction routines, applies diagnosis-conditioned difficulty growth followed by both agent-free and agent-based task validation;
\emph{Training}, which updates the agent via SFT and/or RL;
and \emph{Diagnosis and Feedback}, which analyzes the agent's rollout behaviors to produce a capability analysis report that will be used as feedback for task synthesis in the next iteration.
% No external teacher model, no human-curated task bank, and no learned reward model are required.
Using Qwen3.6-27B as the loop model, \clawloop{} yields consistent improvements on SkillsBench, Claw-Eval-Live, and ClawGym-Bench.
The iteratively refined tasks are further transferred to smaller-scale Qwen3.5-4B and 9B models with gains across all benchmarks, validating data quality independently of the loop model.
% PLACEHOLDER: all numerical claims are synthetic
\end{abstract}

% ======================================================================
%  1  INTRODUCTION
% ======================================================================
\section{Introduction}
\label{sec:intro}

With the rapid advancement of Large Language Models (LLMs),
their deployment scenarios have evolved significantly:
from chatbots to RAG systems, and further from tool-using agents to general-purpose harness-augmented agents~\cite{openclaw2026,openai2025codex,anthropic2025claude37} most recently.
% Inside a harness, an LLM-powered agent operates directly within a user's
% development environment, invoking both pre-defined APIs and
% Turing-complete programming languages (\eg, writing bash scripts,
% composing Python programs, or creating temporary files for intermediate storage).
Unlike prior tool-use agents that interact through predefined tools, harness agents 
are granted with audited and extensible permissions 
and can directly operate on real file systems and shells. 
Such enhanced permission enables LLM-based agents to tackle tasks of substantially greater complexity than tool-use frameworks.

While diverse harnesses have provided a basis for LLMs to take over more workloads from various real-life scenarios, 
these models still require comprehensive post-training targeting harness environments before they can reliably serve as autonomous agents.
The central bottleneck of post-training lies in training data. 
Since harness environments span an enormous task space, 
obtaining diverse training data with high quality has extend far beyond existing data synthesis pipelines, let alone manual annotations. 
% what manual annotation can cover, making automated task synthesis and trajectory collection essential.
While numerous works have proposed various task synthesis pipelines for
general tool-use agents~\cite{song2026envscaler,fang2025towards,cai2025autoforge,tian2026astra,tu2026scaleenv,wang2026agent},
these works are restricted to constructing tasks within fixed domains. 
Specifically, the ``domain'' often consists of several predefined tools with schemas exposed to the agent, 
and the agent's interaction is limited to these tools. 
Such limitation is critical for present-day harness environments, 
% as the agents are exposed to real file systems, command line tools and script languages (\eg, Bash or Python). 
as real-life tasks completed inside a harness generally involve open-ended compositions of file operations, shell commands, skill execution, and multi-step reasoning, 
whose complexity is well beyond predefined tools and cannot be covered by these pipelines. 
% These 
% As such, template-based synthesis that suffices for API-calling agents. 
% preventing LLMs from acquiring the flexible, open-ended
% interaction patterns that harness environments demand.
To the best of our knowledge, ClawGym~\cite{bai2026clawgymscalableframeworkbuilding} is the only work
that performs task synthesis and training inside modern harness environments.
However, it leaves several key gaps unaddressed.
First, existing pipelines assume access to an external {\it stronger model} 
(\eg, MiniMax-M2.5, GLM-5.1)
to provide training signal, 
whether as a task synthesizer,
a trajectory collector, or both. 
The existence of such external models naturally introduces an intrinsic capability ceiling (the capability of the external model itself). 
Second, they are inherently one-shot. 
Tasks are synthesized once based on human prior, the agent is trained once,
and evaluation results inform human researchers rather than being fed back to drive further improvement.

\begin{figure*}[t]
    \centering
    \includegraphics[width=0.85\textwidth]{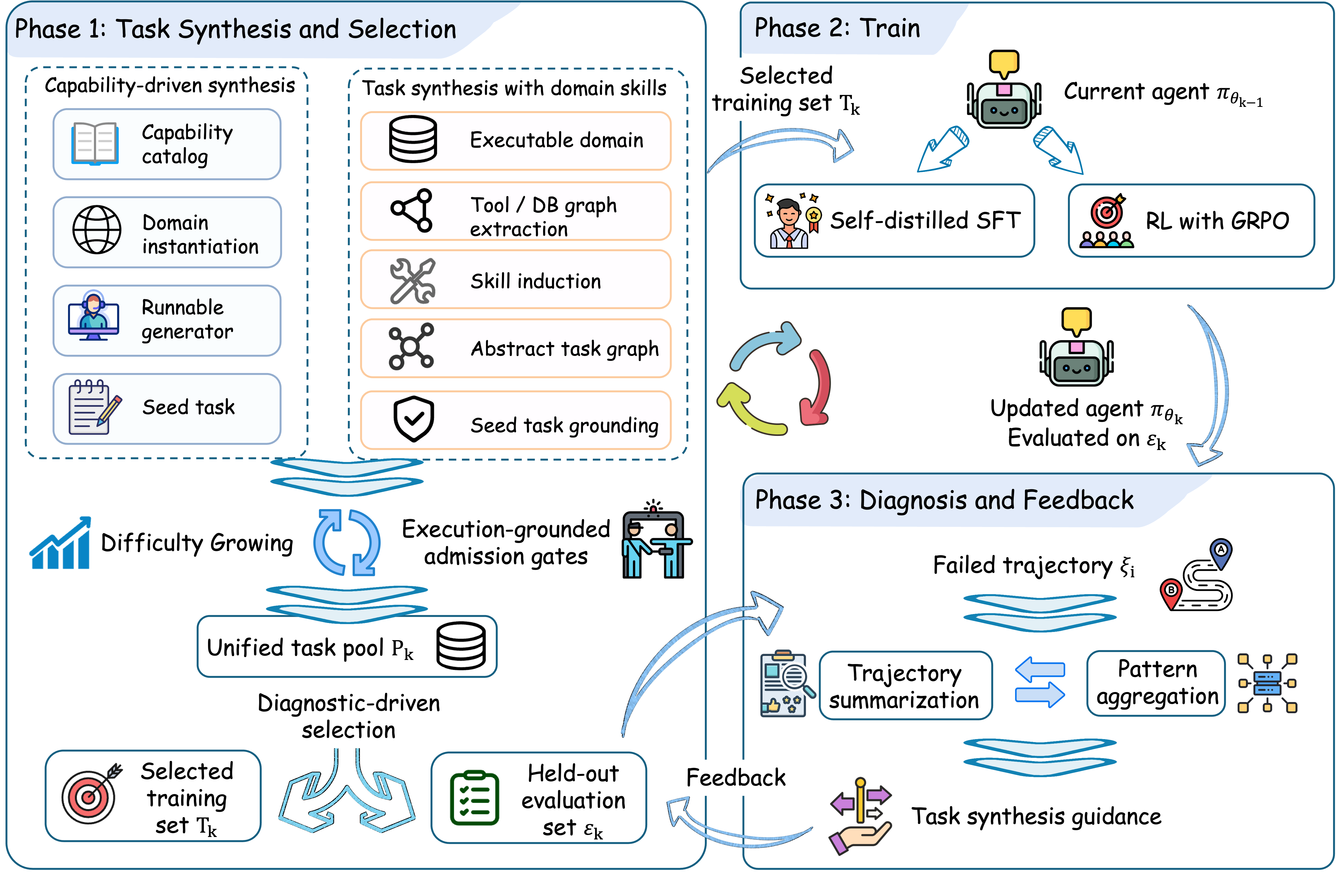}
    \caption{
    Overview of \clawloop{}.
    Within a specific iteration, the agent need to experience three phases: {\it 1. Task Synthesis and Selection, 2. Train} and {\it 3. Diagnosis and Feedback}. 
    All phases operate on the same harness substrate, forming a closed,
    teacher-free self-improvement loop.
    }
    \label{fig:clawloop_overview}
\end{figure*}

We observe that a harness environment already provides the essential
ingredients to close a self-improving loop. 
An LLM model can instantiate arbitrary workspaces, execute agent trajectories,
and verify outcomes via both code-based and rubric-based checks.
What is missing is not infrastructure, but \emph{a mechanism by which the agent
identifies its own weaknesses, generates targeted practice,
and iteratively improves}, 
a capability arguably central to how humans learn.
To this end, we propose \clawloop{},
an iterative self-improving framework that unifies
task synthesis, model training, and self-diagnosis
into a closed loop.Figure~\ref{fig:clawloop_overview} summarizes how these components
interact within one iteration of the closed loop.
% Starting from a modest set of human-provided domain priors (\eg, skill
% definitions and domain keywords), 
\clawloop{} enables the agent to
autonomously synthesize training tasks, collect its own trajectories,
diagnose failure patterns from rollout traces, and generate targeted
tasks for the next round of training, all without an external teacher model.

Our contributions can be summarized as follows:
\begin{itemize}
    \item 
    % \textbf{\clawloop framework.} 
    We propose \clawloop{}, 
    % an entirely \emph{harness-native} and \emph{teacher-free} self-evolving framework for general agents. 
    % Every phase runs inside the same sandboxed environment the agent is trained to operate in, with no external strong model required.
    a closed-loop, teacher-free iterative training framework for general agents. Unlike all prior pipelines that rely on the existence of an external strong model and only operate in a single pass, \clawloop enables the agent to synthesize its own tasks, collect its own trajectories, diagnose its weaknesses, and iteratively improve, all within the same harness environment it is trained to operate in.
    % \item \textbf{Diagnostic-driven task selection.} We introduce a task selection mechanism where the Diagnose phase produces a structured capability profile and uses it to select tasks from a diverse pool that target the agent's weakest categories, creating a targeted training curriculum without requiring curriculum-aware task \emph{synthesis} (\S\ref{sec:taskforge}).
    \item As the core of our proposed \clawloop framework, we design a multi-stage task synthesis mechanism, comprising seed task generation, diagnosis-conditioned difficulty growth, and task validation based on self-reflection. Such mechanism ensures that generated tasks possess rigorous code-level verifiability while maintaining sufficient complexity required for deep reasoning.
    % \item 
    % \textbf{Scalable self-improvement across model sizes.} 
    % The 27B loop driver self-improves via SFT on its own successful trajectories, and the task it produces naturally benefits smaller models (Qwen3.5-4B, Qwen3.5-9B) via RL—all without any teacher model.
    \item 
    % \textbf{Empirical validation with generalization gradient.} 
    Using Qwen3.6-27B inside the loop, we demonstrate that \clawloop yields consistent improvements across diverse benchmarks for harness-based agents. 
    The constructed tasks are also used to train smaller-scale Qwen3.5-4B and 9B models and we observe gains on all benchmarks across different model scales, further validating the quality of these tasks. 
    % To validate the quality of the iteratively refined training data beyond the loop model itself, we further use. 
    Comprehensive ablation studies isolate the contribution of each component and provide a complete understanding of our proposed framework. 
    % The 
    % Across three iterations on three model scales, we demonstrate consistent improvement on three benchmarks spanning a generalization gradient: ClawGym-Bench (within-distribution, as a controlled comparison), SkillsBench~\citep{li2026skillsbench} (cross-domain skill effectiveness across 11 domains), and Claw-Eval-Live~\citep{li2026claw} (live real-world workflows). \clawloop outperforms both strong-teacher and weak-teacher one-shot baselines on all three, confirming that self-improving gains transfer beyond the training distribution.
\end{itemize}

% Unlike ClawGym, which requires external strong models for both task synthesis and trajectory collection and evaluates only on within-distribution tasks, \clawloop's loop is self-bootstrapping and demonstrates gains in \emph{general productivity} that transfer to cross-domain and real-world settings.
% ======================================================================
%  2  RELATED WORK
% ======================================================================
\section{Related Work}
\label{sec:related}

\subsubsection{Task Synthesis and Environment Generation}

The first complete harness-based agent training paradigm is 
ClawGym~\citep{bai2026clawgymscalableframeworkbuilding}, which proposes large-scale synthesis of diverse tasks paired with mock workspaces and hybrid verifiers.
There have also been many works on scalable task synthesis, which can be broadly categorized into three lines: repository-level coding task construction, such as SWE-bench~\citep{jimenez2024swe}, SWE-Gym~\citep{pan2024training}, R2E-Gym~\citep{jain2025r2e}, and SWE-smith~\citep{yang2026swe}; tool-use task and environment synthesis, including ASTRA~\citep{tian2026astra}, AutoForge~\citep{cai2025autoforge}, EnvScaler~\citep{song2026envscaler}, ScaleEnv~\citep{tu2026scaleenv}, Agent World Model~\citep{wang2026agent}, and Agent-World~\citep{dong2026agent}; and capability-oriented task synthesis, such as Self-Challenging~\citep{zhou2026self}, Agent0~\citep{xia2025agent0}, and Tool-R0~\citep{acikgoz2026tool}.These approaches demonstrate the value of executable environments and verifiable tasks, but generally focus on repository-level coding,predefined tool ecosystems, or environment scaling.

\subsubsection{Self-Improving Language Models}

The idea of self-improving in language models has been explored along several axes for many years.
Early approaches bootstrap new supervision from an LLM's own output.
STaR~\citep{zelikman2022star} iteratively trains itself on self-generated
reasoning traces that lead to correct answers, while
Self-Instruct~\citep{wang2023self} expands instruction-tuning data with
model-generated instructions and responses.
The first work that introduces the idea of self-improving to agentic domains is Agent0~\citep{xia2025agent0}, 
which places a Curriculum Agent above an Executor Agent and generates tasks near the executor's capability frontier from zero seed data, retaining tasks by uncertainty, tool-use frequency, and repetition penalties. 
A complementary line keeps the underlying model weights fixed and
treats the non-parametric agent harness as the optimization target.
An example is Agentic Context Engineering
(ACE)~\citep{zhang2025agentic}, 
which evolves a structured context playbook from
successful and failed trajectories.
Self-Harness~\citep{zhang2026self} mines model-specific failure
patterns and converts them into bounded harness modifications that are
accepted only after regression testing.

\subsubsection{Evaluation Benchmarks for LLM-Based Agents with Harness}
The development of various harnesses also calls for comprehensive benchmarks to evaluate the capability of LLM to take real-life workloads with the help of these harnesses. 
The pioneering work, PinchBench~\citep{pinchbench2026}, evaluates models as the reasoning backbone of OpenClaw\cite{openclaw2026} on practical workflows such as research, coding, scheduling, and file management. 
Following it, a growing family of OpenClaw-specific benchmarks has been proposed~\citep{qwenclawbench1.1,ding2026wildclawbench,long2026liveclawbench,zhang2026clawbench,xu2026researchclawbench,meng2026clawmark}. 
% QwenClawBench~\citep{qwenclawbench1.1} focuses on real-user-distribution OpenClaw tasks, WildClawBench~\citep{ding2026wildclawbench} emphasizes bilingual, multimodal, and long-horizon execution in native CLI harnesses, and LiveClawBench~\citep{long2026liveclawbench} evaluates realistic assistant tasks with compositional environment, reasoning, and runtime complexity. 
Among these benchmarks, Claw-Eval~\cite{ye2026clawevaltrustworthyevaluationautonomous} and Claw-Eval-Live~\citep{li2026clawevalliveliveagentbenchmark} evaluate evolving real-world workflows through refreshable task signals and execution-grounded grading. 
ClawGym-Bench~\citep{bai2026clawgymscalableframeworkbuilding} provides executable tasks with mock workspaces and task-specific verifiers, offering a controlled diagnostic setting for Claw-style agents. 
In addition to these OpenClaw-based benchmark, SkillsBench~\citep{li2026skillsbench} is a general agent benchmark that measures whether agents can effectively use procedural Skills across diverse professional domains without being tied to a particular harness, allowing us to assess whether the improvements produced by \clawloop{} generalize beyond OpenClaw-specific environments.

%======================================================================
%  3  PRELIMINARY
% ======================================================================
\section{Preliminary}
\label{sec:preliminary}

We first clarify the boundary of harness in this work. 
Specifically, we use a harness as a sandbox execution environment in which an LLM-based agents can operate.
It generally provides a Docker container with a shell, file system, package manager, and network access, plus APIs for workspace provisioning and trajectory recording.
We use the same harness used for training rollouts and evaluation, ensuring distribution match between training and inference.

% We adopt the task formalism of ClawGym~\citep{bai2026clawgym} with minor notational adjustments.
Based on a given harness, we rigorously define task as a tuple $\tau = \langle p,\, s_0,\, \mathcal{A},\, F,\, V_\tau \rangle$. 
Each component in this tuple is defined as follows:
\begin{itemize}
    \item $p$ is a natural-language instruction that generally describes the task and is exposed to the agent as its initial user prompt;
    \item $s_0 \in \mathcal{S}$ is the initial workspace state (files, environment variables, installed tools), and we use $\mathcal{S}$ to denote the set of all possible workspace states supported by the harness in our loop;
    \item $\mathcal{A}$ is the action space available to the agent (shell commands, file edits, tool invocations);
    \item $F: \mathcal{S} \times \mathcal{A} \to \mathcal{S}$ is the deterministic transition function implemented by the harness sandbox;
    \item $V_\tau: \mathcal{S} \to \{0, 1\}$ is the verifier to check whether an agent has completed the task or not. 
\end{itemize}
We note that such definition is slightly different from general tool-use tasks as in existing literature~\cite{song2026envscaler,tu2026scaleenv,dong2026agent}, 
as the harness introduces additional complexity to both the workspace state (including file systems and installed tools/skills) 
and the agent's action space (including shell scripts or programming languages). 
% \paragraph{Trajectory.}
For a given task $\tau$, an LLM-based agent (denoted as $\pi_\theta$) can then interact with the sandbox environment to produce a trajectory, denoted as $\xi = (s_0, a_1, s_1, \ldots, a_T, s_T)$, where $a_t \sim \pi_\theta(\cdot \mid p, s_0, a_{1:t-1}, s_{1:t-1})$.

% ======================================================================
%  4  CLAWLOOP FRAMEWORK
% ======================================================================
\section{ClawLoop Framework}
\label{sec:framework}

In this section, we describe the complete framework of \clawloop. 
It starts with a pre-trained LLM agent (denoted as $\pi_0$) deployed inside a harness sandbox environment, and consists of $K$ repeating iterations indexed by $k \in \{1, 2, \ldots, K\}$. 
% The core idea is simple: at each iteration, the agent generates its own training curriculum, learns from it, and then diagnoses its remaining weaknesses to inform the next round—forming a closed, self-sustaining improvement loop.
At iteration $k$, based on the most recently updated agent $\pi_{k-1}$, \clawloop proceeds through three phases, as is shown in Figure~\ref{fig:clawloop_overview},:
\begin{enumerate}
    \item \textbf{Task Synthesis and Selection.} The agent first generates a diverse task pool $\mathcal{P}_k$, which is split into a training set $\mathcal{T}_k$ and a held-out evaluation set $\mathcal{E}_k = \mathcal{P}_k \setminus \mathcal{T}_k$.
    Tasks in $\mathcal{T}_k$ are then used to collect rollout trajectories, forming the basis for the next training phase.
    \item \textbf{Training.} Based on the rollout trajectories from $\mathcal{T}_k$, we update the agent $\pi_{\theta_{k-1}} \to \pi_{\theta_k}$ via SFT on successful trajectories and/or RL on tasks that satisfy a reward threshold.
    % \item \textbf{Evaluate} runs $\pi_{\theta_k}$ on the held-out pool split and external benchmarks.
    \item \textbf{Diagnosis and Feedback.} The updated agent $\pi_{\theta_{k}}$ is further evaluated on the held-out evaluation set $\mathcal{E}_k$. The resulting rollout trajectories are fed into the agent itself to identify recurring failure patterns. Any possible feedback will be used in the next iteration to refine the task synthesis and selection phase.
\end{enumerate}

We then describe each phase in detail in the following subsections. 

\subsection{Phase 1: Task Synthesis}
\label{sec:taskforge}

The key component of our framework is task synthesis, which serves as the engine driving iterative self-improvement. Our task synthesis pipeline begins with two complementary routines. The first, \emph{capability-driven synthesis}, starts from a target capability and constructs the workspace environment needed to exercise it.
% generating a task specification, mock project files, and verification criteria around a chosen skill or skill combination. 
The second, \emph{domain-grounded synthesis}, takes the opposite direction. 
It starts from a specific professional domain (\eg database management, workload planning) equipped with domain-specific tools. 
The agent then decomposes it into constituent skills based on its internal structure, and induces concrete tasks from those skills. Together, these two routines produce an initial pool of seed tasks that offer breadth both across capability dimensions and professional domains where domain-specific knowledge is stored in skills. 
The seed tasks are then optionally passed through a \emph{difficulty growth} routine, which adjusts task complexity upward for categories where the agent has already demonstrated proficiency, ensuring that the curriculum remains challenging as the agent improves. 
Finally, all candidate tasks undergo a two-stage validation process—static checks on task well-formedness and workspace integrity, followed by execution-based validation that runs each task's verifier against a gold solution to filter out ill-posed or unsolvable tasks before they enter the training set.

\subsubsection{Capability-driven synthesis}

This route is organized by a catalog of \emph{capabilities} rather than by any concrete toolset.
Each entry binds a workflow pattern---reconciliation, cross-source audit, multi-document merge, or domain calculation---to a target capability, spanning an \textbf{analysis} track whose deliverables are structured artifacts and a \textbf{terminal} track whose deliverables are workspace end-states.
Each capability is then instantiated across a wide range of domains, deliberately including unusual and long-tail ones (\eg beekeeping hive inventories, trail maintenance records, amateur radio contact logs).
Decoupling the capability skeleton from the surface domain in this way---\emph{vary the domain, keep the capability}---lets a single capability be practiced under many different surface forms, which discourages the agent from latching onto domain-specific surface cues instead of the underlying workflow.

The pivotal design decision is that the synthesis LLM never writes task data directly.
Instead it writes a \emph{runnable generator} \verb|build(seed)| $\to (s_0, y, \mathcal{C})$, which \taskforge executes in a sandbox to obtain a multi-source evidence environment $s_0$, the answer key $y$, and an \emph{answer contract} $\mathcal{C}$.
Because $y$ is obtained by executing code over synthesized source data rather than by asking a model to assert an answer, the answer key is grounded in computation, not in model belief.
Generators must be seeded to ensure reproducibility and must create at least two sources that require cross-referencing to complete the task. If generation fails, the sandbox errors are fed back to guide the repair process.
The contract $\mathcal{C}$ addresses a subtler gap: an answer key alone underdetermines a task, since rules used to compute $y$---the filtering criterion, rounding convention, tie-break, or join key---often cannot be recovered from the environment by inspection.
$\mathcal{C}$ therefore declares each such \emph{hidden decision} explicitly, along with a schema fixing the deliverable's shape and numeric tolerance, while containing no answer values.

\subsubsection{Domain-grounded synthesis}

Despite general capabilities that are shared in any domains, 
we further note that real-life tasks often require some {\it domain knowledge} that is unique to a specific domain. 
A simple example may be a term definition, 
while a more complex example may be a complete pipeline to satisfy a requirment. 
Such knowledge may be stored in some {\it skills} for this domain, and an LLM-based agent should be able to quickly grasp and utilize such knowledge to complete domain-specific tasks. 
As such, we need to introduce some professional domains into the task synthesis routine, 
which enable the agent to learn to find, understand and execute such skills to complete tasks within a given domain. 

We first introduce the definition of \emph{executable domain}, 
which is basically a collection of executable tools with schemas exposed to the agent,
together with a database that can completely describe the domain environment and support tool execution. 
Executable domains can be sourced from existing MCP servers
or constructed following existing data synthesis works targeting general tool-user agents~\cite{song2026envscaler, tu2026scaleenv, dong2026agent, wang2026agent}. 
Given an executable domain, we combine abstract syntax tree (AST) inspection with LLM-based analysis (conducted by the agent itself)
to extract a tripartite graph connecting tools,
the atomic parameters and return fields that they consume and produce,
and the database tables that they read and write.
This tripartite graph enables us to cluster tools into \emph{skills}
based on a weighted affinity score over shared graph nodes.
The resulting clusters are provided to the self-evolving agent
as its prior,
which the agent may override when accompanied by a justification.
Each accepted cluster is materialized into
an executable command-line script with database access
rewritten to operate on a state snapshot,
a skill document,
and a capability specification extracted from the generated code
rather than from its documentation.

Tasks are then constructed from the skills available in each domain
through two stages.
In the first stage, an abstract task graph is sampled
according to \emph{structural features}. 
Examples of these features include
long-range dependencies, branching, convergence,
error-recovery paths, and multiple sub-goals. 
These features collectively serve as the difficulty control for this route.
Nodes in this graph represent \emph{knowledge states}
(\eg, ``the order details are now known''), 
and edges between two nodes represent agent actions that transition the workspace from one knowledge state to another. 
The graph contains neither business content nor tool names,
ensuring that its structure is determined before its semantics.

In the second stage, the graph is grounded in a concrete domain.
The initial workspace is seeded with exactly the background
and reference data required to make the task meaningful. 
A coverage check rejects any graph whose transitions
are unsupported by the available skills.
The instruction is then generated under an \emph{information barrier}
that separates it from the gold trajectory,
preventing leakage of the reference solution.
Specifically, the instruction specifies initial parameters
but refers to background records without reproducing their contents,
requiring the agent to retrieve the relevant information;
it defines the goal state while withholding
the procedure, execution order, and branch conditions.
Terminal states are represented as structured predicates
over the final database state
and compiled into an executable test suite.
The tests evaluate only the final state
and locate records using business-semantic keys
rather than hard-coded identifiers,
ensuring that any correct solution can pass.

\subsubsection{Growing Difficulty from Seed Tasks}
% Two independent per-task draws govern difficulty.
While the above two steps can already construct some seed tasks from scratch, 
our pivot studies demonstrate that these tasks often has limited reasoning depth, 
and the agent still fail to overcome real-life task complexity after training with these tasks. 
As such, we consider two approaches for the self-evolving agent to increase task difficulties: {\it multi-hop} and {\it dirtiness} draw. 

The \emph{multi-hop draw} promotes a task to the multi-step retrieval structure typical of realistic agent workloads. Concretely, it produces a workspace with three kinds of sources: an index file that lists entity identifiers and their associated filenames but contains none of the values needed to answer the question; a set of per-entity detail files, no single one of which is sufficient to compute the final answer; and at least one cross-cutting reference source (e.g., a rate table or threshold lookup). Solving such a task requires the agent to read the index, fetch each relevant entity, and join information across all three source types—naturally increasing the number of tool calls and reasoning steps. Importantly, this structure is enforced by construction rather than merely described in the prompt.

The \emph{dirtiness draw} injects controlled noise into the workspace: distractors, missing fields, schema inconsistencies, and unit mismatches. The injection is constrained so that the ground-truth solution logic remains well-defined over the perturbed data, ensuring that tasks stay solvable despite the added friction.

Deliverable formats are round-robined across structured JSON, CSV, multi-file output, and raw reports. For the last case, a machine-checkable sidecar file carries the grading criteria, so that the agent learns to produce free-form text as an output modality while every format remains deterministically verifiable.

\subsubsection{Execution-grounded validation}
% After the task is increased to target difficulty, we further introduce prior validation process that can filter out 
Finally, for a given task, we introduce some validation process to filter out low-quality tasks (\eg, unsolvable tasks or tasks with inconsistency). 
This process involves both agent-free and agent-based validation. 
For agent-free validation, we consider the following aspects: 
% A candidate task must clear four gates, each involving real execution.
\begin{itemize}
    \item \emph{Oracle}: the synthesis LLM writes a reference solution and a deterministic test-based verifier, and the oracle must run and pass its own verifier.
    \item \emph{Leakage audit}: deterministic checks confirm the instruction does not leak the answer key and the verifier admits no trivial solution. Any task whose tests already pass in this state would assign a reward without requiring the agent to perform any work and is therefore discarded as a false positive in reward assignment.
\end{itemize}
For tasks that can pass agent-free validation, we further introduce agent-based validation. 
% which uses the current agent $\pi_{\theta_{k-1}}$ in the loop to try to complete the constructed tasks. 
% To conduct self-distilled SFT on the constructed tasks, 
Specifically, the current agent $\pi_{\theta_{k-1}}$ is deployed in sandbox containers, and its rollouts are collected in parallel across a pool of harness instances for each constructed tasks. 
We let the agent perform $N$ trials for each task, and only keep tasks with number of passed trials in a specific range $[N_l, N_h]$. 
Specifically, we restrict $N_l > 0, N_h < N$, as setting $N_l=0$ will keep tasks that current agent cannot complete (too difficult or with some inconsistency), and setting $N_h = N$ will keep tasks that current agent can always complete (too easy). 

\subsection{Phase 2: Train}
\label{sec:train}

After obtaining a sufficient number of tasks that pass all admission criteria with suitable difficulty levels, \clawloop{} applies standard training techniques to update the agent.
We employ two complementary methods: supervised fine-tuning (SFT) on self-generated successful trajectories, and reinforcement learning (RL) via Group Relative Policy Optimization (GRPO) for further refinement.

\paragraph{Self-distilled SFT with multi-turn loss masking.}
This process directly follows from the task validation step, where each trajectory $\xi_i$ is stored together with its outcome reward $r_i = V_{\tau_i}(s_T^{(i)})$.
Only successful trajectories ($r_i = 1$) for validated tasks are retained for supervised fine-tuning, making this a form of \emph{self-distillation}: the agent learns from its own correct behavior rather than from an external teacher's demonstrations.
% The success rate also provides further. 
% This is the primary training mechanism for the loop-driver model (27B), whose success rate is high enough to yield sufficient positive examples in each iteration.
Following standard practice for multi-turn agent training, we mask the loss on environment observation tokens (tool outputs, file contents) and compute cross-entropy only on the agent's own action tokens:
\begin{equation}
  \mathcal{L}_{\text{SFT}} = -\sum_{i:\, r_i=1}\; \sum_{t=1}^{T_i} m_t^{(i)} \log \pi_\theta\!\bigl(a_t^{(i)} \mid p_i,\, s_0^{(i)},\, a_{1:t-1}^{(i)},\, s_{1:t-1}^{(i)}\bigr),
  \label{eq:sft}
\end{equation}
where $m_t^{(i)} \in \{0,1\}$ is the per-token turn mask that zeroes out all observation positions.
% This masking is critical for harness training: observation tokens often dominate the sequence length (\eg, large file contents or verbose command outputs), and including them in the loss would both dilute the learning signal on agent actions and encourage the model to memorize environment-specific surface patterns.

\paragraph{RL with GRPO.}
While SFT on successful trajectories provides a strong warm-start, it is inherently limited to imitating existing successes and cannot improve beyond the current policy's best behavior.
We therefore apply Group Relative Policy Optimization (GRPO)~\citep{shao2024deepseekmath} with outcome-based rewards to push the agent beyond its SFT ceiling.
For each task $\tau_i$ that passes the validation (neither too easy nor too difficult), the same $N$ rollouts collected during agent-based validation form the group; the advantage of rollout $j$ is computed relative to the group mean:
\begin{equation}
  \hat{A}_j^{(i)} = \frac{r_j^{(i)} - \bar{r}^{(i)}}{\sigma_r^{(i)} + \epsilon}.
  \label{eq:grpo}
\end{equation}

\begin{table}[t]
\centering
\caption{Comparison of different pipelines on Qwen3.6-27B. Best results are \textbf{bolded}.}
\label{tab:loop27b}
\small
\begin{tabular}{@{}lccc@{}}
\toprule
\textbf{Benchmark} & \textbf{Base} & \textbf{ClawGym} & \textbf{ClawLoop} \\
\midrule
SkillsBench (Avg $@$ 5)    & 46.3 & 52.6 & \textbf{57.2} \\
Claw-Eval-Live (Avg Score) & 69.3 & 74.6 & \textbf{76.9} \\
ClawGym-Bench (Avg Score)  & 67.6 & 61.9 & \textbf{68.4} \\
\bottomrule
\end{tabular}
\end{table}

\begin{table*}[t]
\centering
\caption{Performance comparison of Qwen3.5-4B and 9B models trained with different pipelines.}
\label{tab:comparison}
\small
\begin{tabular}{@{}llllccc@{}}
\toprule
\multirow{2}{*}{\textbf{Scale}} & \multirow{2}{*}{\textbf{Model}}  & \multirow{2}{*}{\textbf{Teacher}}  & \textbf{Training} & \textbf{SkillsBench } & \textbf{Claw-Eval-Live } & \textbf{ClawGym-Bench } \\
& & & \textbf{Algorithm} & (Avg $@$ 5) & (Avg Score) & (Avg Score) \\
\midrule
\multirow{3}{*}{4B}
 & Qwen3.5-4B (Base)       & ---  & --- & 5.8 & 32.4 & 33.0 \\ % PLACEHOLDER
 & ClawGym    & MiniMax M2.5+GLM-5.1 & SFT & 6.2 & 36.8 & 34.6 \\
 & ClawLoop  & None         & RL  & \textbf{8.4} & \textbf{48.0} & \textbf{44.1} \\ % PLACEHOLDER
\midrule
\multirow{3}{*}{9B}
 & Qwen3.5-9B (Base)       & ---  & --- & 6.4 & 58.3 & 32.0 \\ % PLACEHOLDER
 & ClawGym & MiniMax M2.5+GLM-5.1 & SFT & 6.7 & 64.3 & 35.3 \\

 & ClawLoop   & None         & RL  & \textbf{10.8} & \textbf{71.7} & \textbf{43.6} \\ % PLACEHOLDER
\bottomrule
\end{tabular}
\end{table*}

\begin{table}[t]
\centering
\caption{Ablation study on different routines used in task synthesis phase for self-improving Qwen3.6-27B models.}
\label{tab:ablation1}
\small
\begin{tabular}{@{}lccc@{}}
\toprule
\textbf{Benchmark} & \textbf{Capability only} & \textbf{Domain only} & \textbf{ClawLoop} \\
\midrule
SkillsBench     & 54.5 & 55.9 & \textbf{57.2} \\
Claw-Eval-Live  & 73.2 & 70.1 & \textbf{76.9} \\
ClawGym-Bench   & 67.2 & 67.8 & \textbf{68.4} \\
\bottomrule
\end{tabular}
\end{table}

\subsection{Phase 3: Diagnosis and Feedback}
\label{sec:diagnose}

To close the self-evolving loop, the agent needs to diagnose its own behavior and provide feedback to the task synthesis phase in next iteration. 
% The basis of this phase is the held-out evaluation set $\mathcal{E}_k$. 
Similar to the validation part in task synthesis, we deploy the updated agent $\pi_{\theta_k}$ in another set of sandbox containers, 
and collect its rollouts for tasks in the held-out evaluation set $\mathcal{E}_k$
In addition to the pass rates on individual tasks, 
\clawloop{} analyzes \emph{why} the agent fails by summarizing failed trajectories, extracting behavioral patterns, and feeding these insights back into the next iteration's task synthesis phase.

\paragraph{Trajectory summarization.}
For each failed trajectory $\xi$ from any tasks in the evaluation set $\mathcal{E}_k$, we prompt the agent itself to produce a structured summary consisting of:
(i)~the \emph{failure point}: the specific action or decision where the trajectory diverged from a successful path;
(ii)~the \emph{root cause}: whether the failure stems from misunderstanding the task, choosing the wrong tool, making incorrect edits, or prematurely terminating;
(iii)~the \emph{missing capability}: what the agent would need to have done differently.
Note that trajectory summarization is a substantially easier task than solving the original problem, as it requires only post-hoc analysis of a concrete execution trace rather than open-ended planning.
These summaries serve as structured diagnostic evidence rather than reward signals.

\paragraph{Pattern aggregation.}
The per-trajectory summaries are then classified by skill category and aggregated to identify recurring \emph{failure patterns}.
Concretely, summaries within each category are clustered by root cause, and the most frequent failure modes are surfaced as diagnostic reports.
For example, consider the capability ``multi-file refactoring'', 
we have observed a common pattern ``agent modifies the target file but fails to update import statements in dependent modules'' during the iterative process. 
These aggregated patterns provide actionable guidance for the next round of task synthesis, directing it to generate tasks that specifically target the identified weaknesses.

% ======================================================================
%  5  EXPERIMENTS
% ======================================================================
\section{Experiments}
\label{sec:experiments}

In this section, we conduct experiments to demonstrate the effectiveness of \clawloop. 
The loop is implemented by OpenClaw~\cite{openclaw2026}, and we use Qwen3.6-27B\cite{qwen3.6-27b} as the agent inside the loop, which runs $K=3$ iterations of the \clawloop cycle, and the task synthesis phase of each iteration generates 1024 new training tasks.
Each task has $N=8$ rollouts for both validation and self-distilled SFT, and we set $N_l=1, N_h=5$ to filter out tasks that are too easy or too hard for RL. 
To evaluate the performance of our trained models, we consider the following three benchmark, covering different specific capabilities of harness-based agents:
\begin{itemize}
    \item Harness-free benchmark. We choose SkillsBench~\citep{li2026skillsbench}, which consists of 86 tasks spanning 11 professional domains. This benchmark targets on evaluating how well agent can utilize skills consisting of domain knowledge, and can be combined to any harness not restricted to OpenClaw. 
% While it is not restricted to any specific harness, it has skills. 
    \item OpenClaw~\cite{openclaw2026}-based benchmark. Among different OpenClaw-based benchmarks, we choose Claw-Eval-Live~\citep{li2026clawevalliveliveagentbenchmark} and ClawGym-Bench~\citep{bai2026clawgymscalableframeworkbuilding} as the representative. Claw-Eval-Live has 105 tasks specifically targeting real-life working scenarios. These tasks do not explicitly depends on specific skills and can evaluate how agent directly working with files. ClawGym-Bench has 200 tasks constructed from the same pipeline to construct training tasks in ClawGym. 
\end{itemize}

\subsection{Main Results}
\label{sec:main_results}

Table~\ref{tab:loop27b} compares \clawloop against the Qwen3.6-27B base model and ClawGym. \clawloop consistently surpasses ClawGym across all three benchmarks by a clear margin, demonstrating that iterative self-improvement with diagnostic feedback produces higher-quality training signal than a single-pass synthesis pipeline. Notably, gains on ClawGym-Bench (within-distribution) are comparable to those on SkillsBench and Claw-Eval-Live, suggesting genuine capability improvement rather than distribution-specific overfitting.

Beyond evaluating the self-improving framework itself, we examine whether the tasks constructed by \clawloop transfer to other models. Smaller models trained on these tasks serve as independent probes of data quality, since any improvement cannot be attributed to the synthesizer's own capacity. Table~\ref{tab:comparison} reports results for Qwen3.5-4B and 9B models trained on \clawloop-generated data versus ClawGym-generated data. Despite ClawGym relying on stronger external teacher models (MiniMax-M2.5 and GLM-5.1) for both task synthesis and trajectory collection, models trained on \clawloop data achieve comparable or superior performance, validating that our iterative, teacher-free pipeline produces training data of equal or higher quality than teacher-dependent alternatives.

\subsection{Ablation Study}
\label{sec:ablation}

\paragraph{Synthesis route}
Note that the two task synthesis routines (capability-driven and domain-grounded) are modularly independent, Table~\ref{tab:ablation1} reports the performance of Qwen3.6-27B within \clawloop when one routine is removed from the task synthesis phase. 
Removing either routine degrades overall performance, but the two routines affect different benchmarks. Dropping the capability-driven routine causes the largest decline on Claw-Eval-Live, whereas dropping the domain-grounded routine primarily hurts SkillsBench and ClawGym-Bench. This asymmetry aligns with the nature of each benchmark: Claw-Eval-Live emphasizes broad, general-purpose agent capabilities that the capability-driven routine is designed to cover, while SkillsBench and ClawGym-Bench are more heavily structured around specific skill categories that the domain-grounded routine targets. 

\begin{figure}[t]
\centering
\includegraphics[width=0.9\columnwidth]{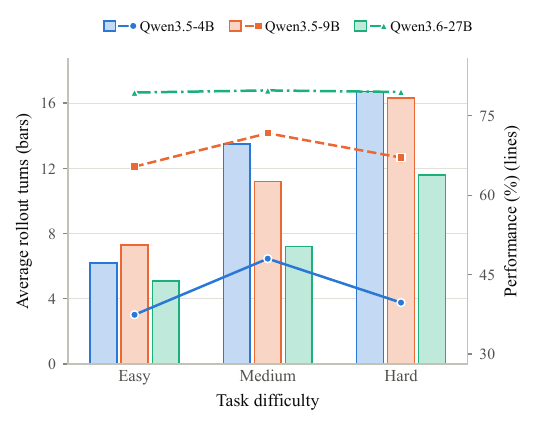}
\caption{Impact on the strength of growing difficulty in task synthesis phase: average rollout turns for constructed tasks and average scores on Claw-Eval-Live.}
\label{fig:difficulty}
\end{figure}

\paragraph{Impact of difficulty growth}
We further analyze the impact of the difficulty growth routine. As shown in Figure~\ref{fig:difficulty}, increasing task difficulty produces an immediate and visible effect: average trajectory length grows substantially, as harder tasks demand more reasoning steps, tool invocations, and error-recovery cycles. However, longer trajectories do not necessarily indicate better training signal. 
In practice, we find that a moderate schedule—raising difficulty by at most one level per iteration—strikes the best balance, yielding trajectories that are challenging enough to drive improvement while maintaining sufficient success rates for effective learning.

\begin{figure}[t]
\centering
\includegraphics[width=0.9\columnwidth]{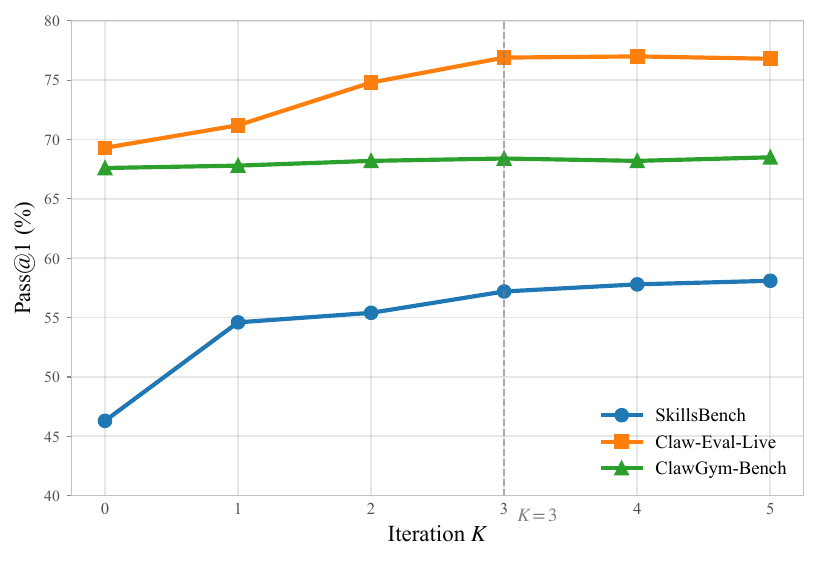}
% \caption{Pass@1 accuracy across \clawloop iterations for all three model scales on SkillsBench (solid) and ClawGym-Bench (dashed). All models show monotonic improvement with diminishing marginal gains.}
\caption{Impact on the number of self-improving iterations $K$ to the model's final performance on different benchmarks}
\label{fig:improvement}
\end{figure}

\paragraph{Impact of $K$.}
Finally, Figure~\ref{fig:improvement} examines how the number of self-improving iterations $K$ affects performance across all three benchmarks. Performance improves steadily in early iterations but saturates at different rates. 
SkillsBench continues to benefit from additional iterations, whereas Claw-Eval-Live and ClawGym-Bench plateau around $K=3$. We attribute this to the broader domain coverage of SkillsBench, which provides a larger space for the curriculum to explore before exhausting novel task diversity. Based on this observation, we set $K=3$ in our main experiments to balance computational cost and final performance.

% ======================================================================
%  7  CONCLUSION
% ======================================================================
\section{Conclusion}
\label{sec:conclusion}

In this paper, we propose \clawloop{}, a self-improving framework for harness-based LLM agents with minimal prior dependencies and no reliance on any external strong teacher model.
As the core component of this self-improving framework, the task synthesis phase starts with two complementary routines that construct seed tasks from different perspectives.
The seed tasks then pass through a diagnosis-conditioned difficulty growth routine and are validated via both agent-free and agent-based procedures to ensure their quality for latter training. 
The complete framework, composing task synthesis, model training, and self-diagnosis into a unified loop, enables agents to identify and address their own weaknesses without human intervention during the training cycle.
Using Qwen3.6-27B inside the loop, we demonstrate that \clawloop{} yields consistent improvements across diverse benchmarks for harness-based agents.
The constructed tasks are further used to train smaller-scale Qwen3.5-4B and 9B models, with gains observed across all benchmarks and model scales.
We leave exploration of longer training horizons with adaptive stopping criteria to future work.

% ======================================================================
%  REFERENCES
% ======================================================================
\bibliography{aaai2027}

\clearpage
\newpage
% ======================================================================
%  APPENDIX
% ======================================================================
\appendix

\section{Experimental Details}
\label{app:details}

This section provides the implementation details omitted from the main text due to space constraints.
We describe the harness and infrastructure configuration (\S\ref{app:infra}), the training
configuration for all model scales (\S\ref{app:trainconfig}), and the evaluation protocol
(\S\ref{app:evalproto}).

\subsection{Harness and Infrastructure}
\label{app:infra}

All three phases of \clawloop{} execute inside the same harness substrate, instantiated
via OpenClaw~\cite{openclaw2026}.
Each rollout runs in an isolated Docker container provisioned from a common base image
that contains a POSIX shell, a Python~3.11 interpreter, standard command-line utilities,
and a package manager.
Containers are network-isolated from external services: any package required by a synthesized
task is pre-installed in the base image or vendored into the workspace at generation time,
which keeps rollouts reproducible and prevents the agent from retrieving solutions from the
open internet.
% PLACEHOLDER: confirm base image contents and Python version
Each container is capped at a wall-clock budget of 24 hours and a maximum of
300 agent turns; exceeding either budget terminates the trajectory and
records it as a failure, so that non-termination is penalized rather than silently discarded.
Rollout collection is parallelized across a pool of 32 concurrent
container instances.

A single \clawloop{} iteration therefore consists of three distinct compute workloads:
task synthesis (LLM inference plus sandboxed generator execution), rollout collection
(LLM inference plus container execution), and gradient updates.
Table~\ref{tab:cost} reports the measured cost of each workload.

\begin{table}[h]
\centering
\caption{Compute cost per \clawloop{} iteration for the Qwen3.6-27B loop model.
All workloads run on the same 32$\times$H800 allocation, so GPU hours are wall-clock
hours scaled by 32.
}
\label{tab:cost}
\small
\begin{tabular}{@{}lcc@{}}
\toprule
\textbf{Workload} & \textbf{Wall-clock (h)} & \textbf{GPU hours} \\
\midrule
Task synthesis (1{,}024 tasks)              & 15.98 & 511 \\
Validation rollouts on $\mathcal{T}_k$      & 30.76 & 984 \\
Training --- SFT                            & 8.03  & 257 \\
Training --- GRPO                           & 18.52 & 593 \\
Diagnosis rollouts on $\mathcal{E}_k$       & 0.94  & 30 \\
\midrule
\textbf{Total per iteration}                & \textbf{74.23} & \textbf{2{,}375} \\
\bottomrule
\end{tabular}
\end{table}

The breakdown shows that rollout collection, not gradient computation, is the dominant
cost: validation rollouts alone account for roughly 41\% of the per-iteration total, and
exceed the combined SFT and GRPO training time.
This is characteristic of harness-native training, where every reward signal requires
executing an agent trajectory against a real environment rather than scoring a static
output.
The diagnosis pass is comparatively cheap because $\mathcal{E}_k$ is a small held-out
split disjoint from the training pool $\mathcal{T}_k$, so only a fraction of the
iteration's tasks are re-run for diagnosis.

\subsection{Training Configuration}
\label{app:trainconfig}

Table~\ref{tab:hyperparams} lists all training hyperparameters.
As described in the main text, \clawloop{} employs two complementary methods within each
iteration: self-distilled SFT on successful trajectories, followed by
GRPO~\citep{shao2024deepseekmath} for further refinement beyond the SFT ceiling.
Every task admitted to the pool receives $N=8$ rollouts, which serve simultaneously as the
validation signal, the source of successful trajectories for SFT, and the group used for
advantage estimation in GRPO.
Reusing a single set of rollouts for all three purposes is what keeps the loop's compute
cost tractable; \S\ref{app:sftrl} examines the consequences for each training stage.

Two details of the SFT setup are specific to the harness setting and worth making explicit.
First, the per-token mask $m_t$ in Eq.~\ref{eq:sft} zeroes out every environment observation
position. This matters more here than in typical multi-turn chat training, because a single
harness observation (a large file read or a verbose build log) can span thousands of tokens
and would otherwise dominate the loss.
Second, trajectories are packed to the maximum sequence length rather than truncated:
truncating a harness trajectory would remove the final verification turns, which are exactly
the behaviors we want the agent to learn.
% Trajectories that exceed the maximum sequence length even after packing are dropped
% (\textbf{[TODO: \%]} of collected trajectories).

\begin{table}[h]
\centering
\caption{Training hyperparameters. Loop-level settings follow the main text;
per-stage optimization settings are listed separately for SFT and GRPO.
% PLACEHOLDER: verify optimizer settings against final run configs
}
\label{tab:hyperparams}
\small
\begin{tabular}{@{}lcc@{}}
\toprule
\textbf{Hyperparameter} & \textbf{SFT} & \textbf{GRPO} \\
\midrule
\multicolumn{3}{@{}l}{\textit{Loop-level settings}} \\
Iterations ($K$)                  & \multicolumn{2}{c}{3} \\
Tasks synthesized per iteration   & \multicolumn{2}{c}{1{,}024} \\
Rollouts per task ($N$)           & \multicolumn{2}{c}{8} \\
Admission band ($N_l$, $N_h$)     & \multicolumn{2}{c}{1, 5} \\
\midrule
\multicolumn{3}{@{}l}{\textit{Optimization settings}} \\
Learning rate                & $2 \times 10^{-5}$   & $5 \times 10^{-7}$   \\  % PLACEHOLDER
Batch size                   & 128                   & 64                    \\  % PLACEHOLDER
Epochs                       & 3                     & 1                     \\  % PLACEHOLDER
Warmup ratio                 & 0.03                  & 0.05                  \\  % PLACEHOLDER
LR schedule                  & cosine                & cosine                \\  % PLACEHOLDER
Max sequence length          & 65{,}536              & 65{,}536              \\  % PLACEHOLDER
KL penalty ($\beta$)         & ---                   & 0.04                  \\  % PLACEHOLDER
Optimizer                    & AdamW                 & AdamW                 \\  % PLACEHOLDER
Weight decay                 & 0.1                   & 0.0                   \\  % PLACEHOLDER
Sampling temperature         & 1.0                   & 1.0                   \\  % PLACEHOLDER
Hardware                     & \multicolumn{2}{c}{32 H800 GPUs}                    \\
\bottomrule
\end{tabular}
\end{table}

\subsection{Evaluation Protocol}
\label{app:evalproto}

All three external benchmarks are evaluated with the harness configuration held fixed
across methods, so that differences reflect the policy rather than the scaffold.
SkillsBench~\citep{li2026skillsbench} is reported as Avg$@$5, \ie the mean over five
independent samples per task, which reduces the variance introduced by its relatively
small size (86 tasks).
Claw-Eval-Live~\citep{li2026clawevalliveliveagentbenchmark} and
ClawGym-Bench~\citep{bai2026clawgymscalableframeworkbuilding} are reported as the average
score produced by their respective official graders.
Decoding uses temperature 0.7 and top-$p$ 0.95 for all evaluations.

None of the three benchmarks is used for training, task selection, or difficulty scheduling
at any point in the loop; the diagnosis phase observes only the held-out synthesized split
$\mathcal{E}_k$.
This separation is what allows the benchmarks to serve as an independent check on whether
the loop produces genuine capability gains rather than fitting its own task distribution.

\section{Complete Algorithm}
\label{app:algorithm}

Algorithm~\ref{alg:clawloop} provides a complete pseudocode description of the \clawloop training loop.

\begin{algorithm}[h]
\caption{\clawloop: Iterative Self-Improvement Loop}
\label{alg:clawloop}
\begin{algorithmic}[1]
\Require Base model $\pi_{\theta_0}$, number of iterations $K$, rollouts per task $N$, admission band $[N_l, N_h]$
\Require Initial diagnostic feedback $\mathcal{F}_0 = \varnothing$
\For{$k = 1$ \textbf{to} $K$}
    \State \textcolor{phase1}{\textbf{// Phase 1: Task Synthesis}}
    \State $\mathcal{P}_k^{\text{raw}} \leftarrow \textsc{DualRouteSynthesis}(\mathcal{F}_{k-1})$ \Comment{Diagnosis-conditioned difficulty growth}
    \State $\mathcal{P}_k^{\text{af}} \leftarrow \textsc{AgentFreeValidation}(\mathcal{P}_k^{\text{raw}})$ \Comment{Generator runs, verifier executable}
    \For{each task $\tau_i \in \mathcal{P}_k^{\text{af}}$} \Comment{Agent-based validation}
        \For{$j = 1$ \textbf{to} $N$}
            \State $\xi_j^{(i)} \leftarrow \text{Rollout}(\pi_{\theta_{k-1}}, \tau_i)$ in harness sandbox
            \State $r_j^{(i)} \leftarrow V_{\tau_i}(s_T^{(i)})$
        \EndFor
    \EndFor
    \State $\mathcal{P}_k \leftarrow \{\tau_i : N_l \leq \sum_j r_j^{(i)} \leq N_h\}$ \Comment{Neither too easy nor too hard}
    \State $\mathcal{T}_k,\, \mathcal{E}_k \leftarrow \textsc{Split}(\mathcal{P}_k)$ \Comment{$\mathcal{E}_k = \mathcal{P}_k \setminus \mathcal{T}_k$}
    \State
    \State \textcolor{phase2}{\textbf{// Phase 2: Train}}
    \State $\mathcal{D}_{\text{SFT}} \leftarrow \{\xi_j^{(i)} : \tau_i \in \mathcal{T}_k,\ r_j^{(i)} = 1\}$ \Comment{Reuse validation rollouts}
    \State $\theta' \leftarrow \text{SFT}(\pi_{\theta_{k-1}},\, \mathcal{D}_{\text{SFT}})$ \Comment{Multi-turn loss masking}
    \State $\theta_k \leftarrow \text{GRPO}(\pi_{\theta'},\, \{(\tau_i, \{\xi_j^{(i)}\}_{j=1}^{N})\}_{\tau_i \in \mathcal{T}_k})$
    \State
    \State \textcolor{phase3}{\textbf{// Phase 3: Diagnose and Feedback}}
    \State $\Xi_k \leftarrow \{\text{Rollout}(\pi_{\theta_k}, \tau) : \tau \in \mathcal{E}_k\}$ \Comment{Held-out split only}
    \State $\text{summaries}_k \leftarrow \textsc{SummarizeFailures}(\{\xi \in \Xi_k : r(\xi) = 0\})$
    \Statex \hfill \Comment{Failure point, root cause, missing capability}
    \State $\mathcal{F}_k \leftarrow \textsc{AggregatePatterns}(\text{summaries}_k)$ \Comment{Recurring failure patterns}
\EndFor
\State \Return $\pi_{\theta_K}$
\end{algorithmic}
\end{algorithm}

\section{SFT and RL within the Loop}
\label{app:sftrl}

The main text describes SFT and GRPO as complementary stages: SFT provides a warm-start by
imitating the agent's own successes, and GRPO pushes past the ceiling that imitation imposes.
This section explains why the two stages contribute differently at different model scales.

\paragraph{Why the two stages contribute differently.}
Self-distilled SFT requires a supply of successful trajectories, and that supply is a
function of the model's own success rate on the validated pool.
The 27B loop model clears the admission band on enough tasks that each iteration yields
between 892 and 1{,}978 successful trajectories (Table~\ref{tab:loopstats}), which is
sufficient for SFT alone to produce a stable gradient signal.
Smaller models do not: on the same pool their success rate is low enough that SFT sees few
positive examples, and the resulting update is both noisy and biased toward the easiest
tasks in the pool.
GRPO compensates because it extracts signal from the \emph{contrast} between rollouts within
a group, so a task on which the model succeeds once out of eight attempts still produces a
usable advantage estimate, whereas SFT would retain that single trajectory and discard the
seven informative failures.
This is why the gains reported for the 4B and 9B models in Table~\ref{tab:comparison} are
attributed primarily to the RL stage.

\paragraph{Interaction with task admission.}
This also explains why the admission band $[N_l, N_h]$ with $N_l > 0$ and $N_h < N$ matters
more for RL than for SFT.
Tasks solved on every attempt produce zero advantage under Eq.~\ref{eq:grpo} because the
group is unanimous, and tasks solved on no attempt produce zero advantage for the same
reason; in both cases the group standard deviation $\sigma_r$ collapses and the task
contributes nothing but compute.
The main text's setting of $N_l = 1$ and $N_h = 5$ out of $N = 8$ rollouts retains exactly
the tasks whose outcome is uncertain under the current policy, which is where the gradient
signal is concentrated.
Because the band is defined relative to $\pi_{\theta_{k-1}}$, it moves with the agent:
a task that is informative at iteration $k$ may be unanimously solved by iteration $k+1$
and is then excluded automatically, without any external curriculum schedule.

\paragraph{Scaling behavior.}
The main text reports \clawloop{} results at three scales: Qwen3.5-4B and 9B as consumers of
the curriculum, and Qwen3.6-27B as the loop model.
Two caveats bound what these results establish about scaling.
First, a single loop model drives the curriculum at all three scales, so we cannot separate
the effect of scale on the \emph{consumer} of the curriculum from its effect on the
\emph{producer}; varying the loop model itself would be required for that.
Second, the 4B and 9B models never run their own diagnosis phase, and instead inherit a
curriculum shaped by the 27B model's weakness profile.
That this transfers at all is precisely the finding reported in the main text, and it is
what licenses treating the smaller models as independent probes of data quality.
But it also means those models are not self-improving in the sense the framework defines,
and their gains should be read as evidence about the curriculum rather than as instances of
the full loop.

\section{Loop Dynamics}
\label{app:dynamics}

The main text reports what the loop achieves at its endpoint.
This section reports what happens \emph{inside} the loop: how the task pool and the
trajectories it elicits evolve across iterations.
These quantities are the loop's internal state, and tracking them is what allows us to
distinguish genuine curriculum progression from a pool that merely grows without becoming
more informative.

\subsection{Per-Iteration Statistics}

Table~\ref{tab:loopstats} summarizes the state of the loop at each iteration.
The two validation stages play distinct roles.
Agent-free validation checks only that a candidate is well-formed --- that its generator
runs deterministically and its verifier is executable --- and it admits a stable
$\sim$89\% of candidates in every iteration, indicating that the synthesis routine's basic
output quality does not degrade as the loop proceeds.
Agent-based validation is the selective stage: it retains only tasks whose difficulty falls
inside the admission band $[N_l, N_h] = [1, 5]$, and because that band is defined relative
to the current policy, it is the stage through which the agent's own improvement feeds back
into the curriculum.

\begin{table}[h]
\centering
\caption{Per-iteration loop statistics for the Qwen3.6-27B loop model.
The admission rate is reported relative to the 1{,}024 candidates synthesized per
iteration.
}
\label{tab:loopstats}
\small
\begin{tabular}{@{}lccc@{}}
\toprule
\textbf{Quantity} & \textbf{$k{=}1$} & \textbf{$k{=}2$} & \textbf{$k{=}3$} \\
\midrule
Candidates synthesized       & \multicolumn{3}{c}{1{,}024} \\
Passed agent-free validation & 897 & 910 & 913 \\
Passed agent-based validation & 672 & 469 & 567 \\
Admission rate (\%)          & 65.6 & 45.8 & 55.4 \\
\midrule
SFT trajectories retained    & 1{,}978 & 892 & 1{,}694 \\
Mean trajectory turns        & 5.6 & 11.4 & 8.7 \\
Mean trajectory tokens       & 1{,}207 & 2{,}946 & 1{,}899 \\
\bottomrule
\end{tabular}
\end{table}

The admission rate does not decay monotonically: it falls sharply from 65.6\% at $k{=}1$ to
45.8\% at $k{=}2$, then recovers to 55.4\% at $k{=}3$.
The trajectory statistics move in lockstep, with mean turns rising from 5.6 to 11.4 before
settling at 8.7 and mean token counts tracing the same shape.
This is the difficulty-growth routine overshooting and then correcting.
At $k{=}2$ the routine escalates difficulty too aggressively in response to the agent's
first-round gains, pushing many candidates past the upper bound $N_h$: the admission rate
drops, and the tasks that do survive are markedly longer.
By $k{=}3$ the escalation has been damped and the pool settles at a difficulty the agent
can engage productively, with the admission rate and trajectory length both landing between
their $k{=}1$ and $k{=}2$ values.
The loop therefore converges on an appropriate difficulty level by traversing it from both
sides rather than approaching it monotonically.

\subsection{Diminishing Returns}

Figure~\ref{fig:improvement} in the main text shows that performance saturates around
$K=3$ for two of the three benchmarks.
Table~\ref{tab:loopstats} suggests a mechanistic reading.
The binding constraint is not how many candidates are synthesized, which is fixed at
1{,}024 per iteration, but how many land inside the admission band: only 45.8\% to 65.6\%
survive agent-based validation, so between a third and a half of the synthesis budget
yields no training signal at all.
Discarded tasks are those the agent solves on every attempt or on none, and both produce
zero advantage under Eq.~\ref{eq:grpo} because the group is unanimous.
The $k{=}2$ overshoot shows that keeping candidates inside the band requires the synthesis
routine to track a moving policy, and that this tracking is imprecise.
This points to the difficulty controller, rather than the number of iterations, as the
limiting factor: a routine that placed a larger fraction of its output inside the band
would extract more from the same budget.

\section{Case Study}
\label{app:casestudy}

To make the synthesized tasks concrete, this section examines two instances drawn from the
same task family in the \taskforge{} pool, together with the trajectories they elicited from
the loop model.
The two instances differ substantially in difficulty, and comparing them shows \emph{where}
the additional difficulty is absorbed: not in a longer sequence of task steps, but in a
markedly longer verification phase.
Task instructions are translated from the original Chinese.

\subsection{Two Instances of an Aggregation Task}

Both instances belong to an aggregation-and-reporting family in which the agent must join
two CSV inputs, compute grouped statistics under explicit business rules, and emit report
files that a verifier checks field by field.

\paragraph{Instance A (harder).}
\begin{quote}\small\itshape
From the order ledger (\texttt{orders.csv}) and the product master table
(\texttt{products.csv}), generate a group of order-amount distribution and category-tier
reports (covering distribution statistics, quartile analysis, refund metrics, value-tier
summaries, and a voided-order audit log), writing multiple JSON files to \texttt{/output/}.
Detailed requirements are in \texttt{/input/requirements.md}; the output contract is in
\texttt{/input/task\_spec.json}.
\end{quote}

\paragraph{Instance B (easier).}
\begin{quote}\small\itshape
Read \texttt{/input/orders.csv} and \texttt{/input/customers.csv} and join orders to
customers by \texttt{customer\_id}. For each customer compute a spending profile
(\texttt{total\_paid}, \texttt{total\_refunded}, \texttt{net\_spending}, and the
corresponding order counts), assign a tier by \texttt{net\_spending}
(\texttt{HIGH\_VALUE} $\geq 500$, \texttt{STANDARD} for $0 < \cdot < 500$,
\texttt{AT\_RISK} $\leq 0$), and write each tier to its own CSV file plus a
\texttt{tier\_summary.csv} aggregate.
\end{quote}

\paragraph{Workspace and hidden decisions.}
Neither instance is answerable from any single input file: the order ledger carries
transaction amounts but not the category or customer attributes needed to group them, and
the joined dimension table supplies those attributes without the amounts.
Both instances also depend on decisions that the data alone underdetermines, and these are
declared in skill files that the workspace makes available to the agent rather than left
implicit.
For Instance A, the tier-classification skill specifies that a category median falling
exactly on a threshold is assigned to the higher tier, that tiers are computed from the
unrounded median with rounding applied only at output, and that categories with no valid
orders are omitted from the summary entirely.
Instance B is governed analogously by skills covering decimal money aggregation, dirty-record
handling, and the representation of empty tiers.
Without such declarations the tasks would be ill-posed: several defensible outputs would
exist, and a verifier checking exact values would penalize the agent for a choice the
instruction never constrained.

\subsection{The Two Trajectories}

Table~\ref{tab:casetrace} places the two trajectories side by side, grouped into phases.
Both agents solved their instance, and both were admitted to the training pool.

\begin{table}[h]
\centering
\caption{The two trajectories by phase. Turn indices refer to agent turns; a single turn
may issue several tool calls.}
\label{tab:casetrace}
\small
\begin{tabular}{@{}lp{0.34\columnwidth}p{0.34\columnwidth}@{}}
\toprule
\textbf{Phase} & \textbf{Instance A (harder)} & \textbf{Instance B (easier)} \\
\midrule
Explore
& A1--A5: read requirements, output contract, both CSVs, config, and six skill files
& A1: read both CSVs and five skill files \\
\addlinespace
Implement
& A6: write generation script
& A2--A3: create output directory, write script \\
\addlinespace
Execute
& A7: run script
& A4: run script \\
\addlinespace
Verify
& A8--A13: recompute aggregates independently, dump and re-read all three outputs, cross-check tier totals against the distribution report, audit sort order
& A5: read back all four outputs \\
\addlinespace
Finalize
& A14--A15: remove scratch script, report
& A6: report \\
\bottomrule
\end{tabular}
\end{table}

The phase structure is identical, and the two trajectories diverge almost entirely in the
verification phase.
Instance A devotes six of its fifteen turns (40\%) to checking its own output, whereas
Instance B spends one turn of six (17\%).
Table~\ref{tab:casemetrics} quantifies the gap.

\begin{table}[h]
\centering
\caption{Trajectory statistics for the two instances.}
\label{tab:casemetrics}
\small
\begin{tabular}{@{}lcc@{}}
\toprule
\textbf{Quantity} & \textbf{Instance A} & \textbf{Instance B} \\
\midrule
Difficulty score        & 0.53 & 0.14 \\
\midrule
Agent turns             & 15   & 6 \\
Tool calls              & 24   & 14 \\
Shell executions        & 9    & 2 \\
Skill files consulted   & 6    & 5 \\
Reasoning tokens        & 67{,}994 & 4{,}067 \\
\bottomrule
\end{tabular}
\end{table}

The scaling is uneven, and the pattern is informative.
Turn count grows by $2.5\times$ and tool calls by only $1.7\times$, but shell executions
grow by $4.5\times$ and reasoning tokens by nearly $17\times$.
The harder instance does not require many more \emph{actions}; it requires far more
deliberation per action, and most of that deliberation is spent confirming that the actions
already taken were correct.

\subsection{What the Verification Phase Contains}

The content of Instance A's verification turns is worth describing, because it is the
behavior that self-distilled SFT retains when the trajectory is admitted.
After running its script, the agent recomputed the global revenue aggregate independently
and obtained a figure that appeared to contradict its own earlier estimate; it then
re-derived the sum term by term, found the true total to be $6984.997$, and confirmed that
this rounds to the reported $6985.00$.
It next cross-checked the tier summary against the distribution report, verifying that the
per-category revenues sum to the reported tier totals.
Finally, its audit of sort order flagged an apparent violation among categories sharing an
identical median; on inspection the agent determined that its own checking script was at
fault, having rejected equal medians outright instead of verifying that the alphabetical
tie-break had been applied correctly, and that the output was in fact correct.

This last step is the most interesting of the three, because the agent is not merely
re-running its solution but auditing its own audit, and correctly concluding that the
solution stands and the check was wrong.
Behavior of this kind is what the admission band is designed to select for.
A task that the agent solves on the first attempt without verification produces a trajectory
with little of this content, while a task it cannot solve at all produces none; tasks in
between elicit trajectories in which the agent must check itself, and those are the
trajectories that enter the SFT set.

We note that these two instances are illustrative rather than a controlled comparison.
They come from the same task family but are not the same task, so the difference in
verification effort cannot be attributed to difficulty alone with any precision.
The aggregate trend in Table~\ref{tab:loopstats}, where mean trajectory length tracks the
admission rate across iterations, is the quantitative form of the same observation.

\end{document}